\pdfoutput=1
\documentclass[11pt]{article}

\usepackage{acl}

\usepackage{times}
\usepackage{latexsym}

\usepackage{xspace}
\usepackage{amsmath}
\usepackage{subcaption}
\usepackage{tabularx}
\usepackage{booktabs}

\usepackage{graphicx}
\usepackage{enumitem}
\usepackage{csvsimple}
\usepackage{dsfont}
\usepackage{pifont}
\usepackage{amsthm}
\usepackage[ruled,vlined]{algorithm2e}
\usepackage{xspace}
\usepackage{amsmath}
\usepackage{arydshln}
\usepackage{multirow}
\usepackage{colortbl}
\usepackage{float}
\usepackage{multicol}
\usepackage{soul}
\usepackage{pifont}
\usepackage{array,booktabs,ragged2e}
\usepackage{colortbl}
\usepackage[normalem]{ulem} % for \sout

\definecolor{brightmaroon}{rgb}{0.76, 0.23, 0.28}

\newcolumntype{P}[1]{>{\centering\arraybackslash}p{#1}}

\newcommand{\scifact}{\textsc{SciFact}\xspace}
\newcommand{\sciver}{\textsc{SciVer}\xspace}

\newcommand{\ts}{\: \texttt{<s>} \:}
\newcommand{\tsl}{\: \texttt{</s>} \:}
\newcommand{\tslsub}[1]{\: \texttt{</s>}_{#1} \:}

\newcommand{\lamRationale}{\lambda_{\textrm{rationale}}}

\newcommand{\fever}{\textsc{Fever}\xspace}
\newcommand{\pqaa}{\textsc{PQA-A}\xspace}
\newcommand{\pqa}{\textsc{PubmedQA}\xspace}
\newcommand{\ei}{\textsc{EvidenceInference}\xspace}
\newcommand{\storc}{\textsc{S2ORC}\xspace}
\newcommand{\climatefever}{\textsc{Climate-Fever}\xspace}
\newcommand{\covidFact}{COVIDFact\xspace}
\newcommand{\healthVer}{HealthVer\xspace}
\newcommand{\pubhealth}{\textsc{PubHealth}\xspace}

\newcommand{\bert}{BERT\xspace}
\newcommand{\roberta}{RoBERTa\xspace}

\newcommand{\longformer}{Longformer\xspace}
\newcommand{\pjoint}{\textsc{ParagraphJoint}\xspace}
\newcommand{\arsjoint}{\textsc{ARSJoint}\xspace}
\newcommand{\sysname}{\textsc{MultiVerS}\xspace}

\newcommand{\vertserini}{\textsc{Vert5Erini}\xspace}

\newcommand{\feverSci}{\textsc{FeverSci}\xspace}
\newcommand{\targetOnly}{No-Pretrain\xspace}

\newcommand{\nei}{\textsc{NEI}\xspace}
\newcommand{\support}{\textsc{Support}\xspace}
\newcommand{\refute}{\textsc{Refute}\xspace}
\newcommand{\supports}{\textsc{Supports}\xspace}
\newcommand{\refutes}{\textsc{Refutes}\xspace}

\newcommand{\refuted}{\textsc{Refuted}\xspace}
\newcommand{\yhat}{\widehat{y}}
\newcommand{\rhat}{\widehat{r}}
\newcommand{\Rhat}{\widehat{R}}

\newcommand{\numHumanAnnots}{151\xspace}

\newcommand{\cmark}{\ding{51}\xspace}%
\newcommand{\xmark}{\ding{55}\xspace}%

\newcommand{\sz}[1]{\scriptsize #1}

\definecolor{lightpink}{rgb}{1.0, 0.71, 0.76}
\definecolor{lightblue}{RGB}{212, 235, 255}
\definecolor{salmon}{RGB}{255, 164, 168}
\definecolor{lightgreen}{RGB}{177, 231, 171}
\definecolor{lightyellow}{RGB}{255, 255, 148}
\sethlcolor{lightblue}

\newcommand\hlc[2]{\sethlcolor{#1}\hl{#2}}

\newcolumntype{Y}{>{\centering\arraybackslash}X}
\newcolumntype{s}{>{\hsize=.5\hsize}X}
\newcolumntype{L}[1]{>{\raggedright\let\newline\\\arraybackslash\hspace{0pt}}m{#1}}
\newcolumntype{R}[1]{>{\RaggedLeft\arraybackslash}p{#1}}
\newcommand\tworows[1]{\multirow{2}{*}{\shortstack[l]{#1}}}
\newcommand\tworowsc[1]{\multirow{2}{*}{\shortstack[c]{#1}}}

\usepackage[T1]{fontenc}
\usepackage[utf8]{inputenc}

\usepackage{microtype}

\title{\sysname: Improving scientific claim verification with \\ weak supervision and full-document context}
\author{David Wadden$^\mathbf{\dagger}$ \quad
  Kyle Lo$^\mathbf{\ddagger}$ \quad
  Lucy Lu Wang$^\mathbf{\ddagger}$ \\
  {\bf Arman Cohan}$^\mathbf{\ddagger}$ \quad
  {\bf Iz Beltagy}$^\mathbf{\ddagger}$ \quad
  {\bf Hannaneh Hajishirzi}$^\mathbf{\dagger\ddagger}$\\
  $^\mathbf{\dagger}$
  University of Washington, Seattle, WA, USA \\
  $^\mathbf{\ddagger}$
  Allen Institute for Artificial Intelligence, Seattle, WA, USA \\
  {\tt\small \{dwadden,hannaneh\}@cs.washington.edu} \\
  {\tt\small \{kylel,lucyw,armanc,beltagy\}@allenai.org}
}

\begin{document}
\maketitle

\begin{abstract}
    The scientific claim verification task requires an NLP system to label scientific documents which \support or \refute an input claim, and to select evidentiary sentences (or \emph{rationales}) justifying each predicted label. In this work, we present \sysname, which predicts a fact-checking label and identifies rationales in a multitask fashion based on a shared encoding of the claim and full document context. This approach accomplishes two key modeling goals. First, it ensures that all relevant contextual information is incorporated into each labeling decision. Second, it enables the model to learn from instances annotated with a document-level fact-checking label, but lacking sentence-level rationales. This allows \sysname to perform weakly-supervised domain adaptation by training on scientific documents labeled using high-precision heuristics. Our approach outperforms two competitive baselines on three scientific claim verification datasets, with particularly strong performance in zero / few-shot domain adaptation experiments. Our code and data are available at \url{https://github.com/dwadden/multivers}.
\end{abstract}

\section{Introduction} \label{sec:intro}

\begin{figure}[t]
  \centering
  \includegraphics[scale=0.7]{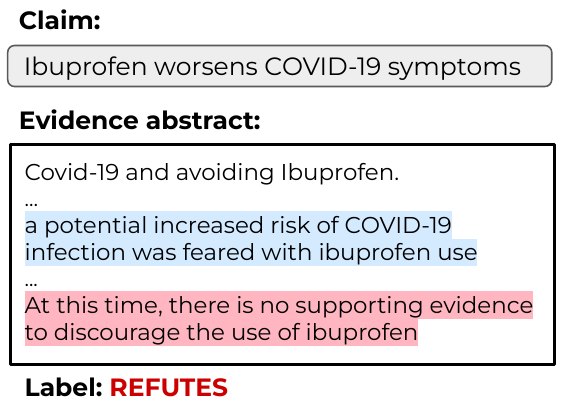}
  \caption{A claim from the \healthVer dataset, refuted by a research abstract. The sentence in \hlc{lightpink}{red} is a \emph{rationale} reporting a finding that \refutes the claim. However, this finding cannot be interpreted properly without the context in \hlc{lightblue}{blue}, which specifies that the finding applies to Ibuprofen as a treatment for COVID. \sysname incorporates the full context of the evidence-containing abstract when predicting fact-checking labels.}
  \label{fig:teaser}
\end{figure}

The proliferation of scientific mis- and dis-information on the web has motivated the release of a number of new datasets \cite{Saakyan2021COVIDFactFE,Sarrouti2021EvidencebasedFO,Wadden2020FactOF,Kotonya2020ExplainableAF} and the development of modeling approaches \cite{Pradeep2021ScientificCV,Li2021APM,Zhang2021AbstractRS} for the task of \emph{scientific claim verification}. The goal of the task is to verify a given scientific claim by labeling scientific research abstracts which \support or \refute the claim, and to select evidentiary sentences (or \emph{rationales}) reporting the findings which justify each label.

% Data collection for new scientific domains is challenging since it requires  expert annotators with domain knowledge. 

A common approach to this task is to first extract rationales from the larger document context, and then make label predictions conditioned on the selected rationales. 
This ``extract-then-label'' approach has two important drawbacks, which we aim to address in this work. First, the rationales may lack information required to make a prediction when taken out-of-context; for instance, they may contain acronyms or unresolved coreferences, or lack qualifiers that specify the scope of a reported finding (Figure \ref{fig:teaser} provides an example). Second, the ``extract-then-label'' approach requires training data annotated with both sentence-level rationales and abstract-level labels. While sentence-level rationale annotations are costly and require trained experts, abstract-level labels can be created cheaply using high-precision heuristics, e.g., the titles of research papers sometimes make claims that are supported by their abstracts.

% Second, the ``extract-then-label'' approach is unable to take advantage of abstract-level labels without sentence-level rationale annotations. This presents a bottleneck for the development of fact verification models in specialized domains: rationale annotations are difficult and time-consuming to produce, but we demonstrate in this work that abstract-level labels can be derived cheaply using high-precision weak supervision heuristics, enabling domain adaptation without extensive manual annotation.

Motivated by these challenges, we introduce \sysname (\textbf{Multi}task \textbf{Ver}ification for \textbf{S}cience): Given a claim and evidence-containing scientific abstract, \sysname creates a shared encoding of the entire claim / abstract context, using the \longformer encoder \cite{Beltagy2020LongformerTL} to accommodate long sequences. Then, it predicts an abstract-level fact-checking label and sentence-level rationales in a multitask fashion, enforcing consistency between the outputs of the two tasks during decoding. This modeling approach ensures that label predictions are made based on all available context, and enables training on instances derived via weak supervision for which abstract-level labels are available, but sentence-level rationales are not. 

In experiments on three scientific claim verification datasets, we find that \sysname outperforms two state-of-the-art baselines, one of which has more than 10x the parameters of our system. In addition, we show that training \sysname on weakly-labeled in-domain data substantially improves performance in the zero / few-shot domain adaptation settings. The ability to achieve reasonable performance given limited labeled data is especially valuable in specialized domains, due to the high cost of collecting expert annotations.

% I feel like adding the unified thing doesn't quite work.

% We perform experiments on three scientific claim verification datasets focusing on the biomedical domain.
% % in different formats. 
% We convert all datasets to a unified format, pairing claims with full scientific abstracts. In addition, we apply weak supervision heuristics to convert two large scientific datasets not created for claim verification to the same format, and use them as additional weakly-labeled training data.
% % In experiments on three scientific claim verification datasets, 

% We find that \sysname outperforms two state-of-the-art baselines, one of which has more than 10x the parameters of our system. Training on weakly-labeled in-domain data is especially beneficial for zero / few-shot domain adaptation, which points toward the potential of \sysname to enable domain adaptation for sub-fields -- in science and beyond -- where annotations are expensive.

In summary, our contributions are as follows:

\begin{enumerate}[noitemsep, topsep=1pt, leftmargin=*]
    \item We introduce \sysname, a multitask system for full-context scientific claim verification. \sysname improves fully-supervised fact-verification performance by an average of 11\% on three datasets over two state-of-the-art baselines, with improvements of 14\% and 26\% in the few-shot and zero-shot settings.
    \item We present weak supervision heuristics to assign fact-checking labels to two large scientific datasets, and show that training on these annotations more than doubles zero-shot domain adaptation performance.
    \item  Through ablations and analysis, we demonstrate that our multitask modeling approach achieves our goals of incorporating full-document context into label predictions, and facilitating zero / few-shot domain adaptation.
\end{enumerate}

\section{Background} \label{sec:background}

\subsection{The scientific claim verification task} \label{sec:task_review}

We use the definition of scientific claim verification from the \scifact task \cite{Wadden2020FactOF}, and provide a brief overview of the task here.
% and refer the reader to that work for more detail. 
Other works have cast scientific claim verification as a sentence-level natural language inference (NLI) task; in \S \ref{sec:target_data}, we describe how we process these datasets to be compatible with the task as considered in this work. 

\paragraph{Task definition} Given a claim $c$ and a collection of \emph{candidate abstracts} which may contain evidence relevant to $c$, the scientific claim verification task requires a system to predict a \emph{label} $y(c, a) \in \{\supports, \refutes, \nei\footnote{\nei stands for ``Not Enough Info''.}\}$, which indicates the relationship between $c$ and $a$ for each candidate $a$. For all abstracts labeled \supports or \refutes, the system must also identify \emph{rationales} $R(c, a) = \{ r_1(c, a), \dots, r_n(c, a) \}$, where each $r_i(c, a)$ is a sentence from $a$ that either entails or contradicts the label $y(c, a)$.\footnote{This rationale definition is simplified slightly from the one presented in \citet{Wadden2020FactOF}.} The rationales may not be self-contained, and may require additional context from elsewhere in the abstract to resolve coreferential expressions or acronyms, or to determine qualifiers specifying experimental context or study population.\footnote{This convention is consistent with related tasks in rationalized NLP for biomedical literature, such as \citet{Lehman2019InferringWM} and \citet{DeYoung2020EvidenceI2}.} Examples of these situations are provided in Figure \ref{fig:teaser} and Appendix \ref{appx:context_dependent}.

\paragraph{Evaluation}

The \scifact task reports four evaluation metrics. We have found that two of these metrics are sufficient to convey the important findings for our experiments: (1) \emph{abstract-level label-only} evaluation computes the model's F1 score in identifying abstracts that \support or \refute each claim. Predicting the correct label $y(c, a)$ is sufficient; models do not need to provide rationales. (2) \emph{Sentence-level selection+label} evaluation computes the point-wise product of the model's F1 score in identifying the rationales $R(c, a)$, with the model's abstract-level label $y(c, a)$; this metric rewards precision in identifying exactly which sentences contain the evidence justifying the label. In this work, we refer to these two metrics as ``abstract'' and ``sentence'' evaluation respectively.

\paragraph{Retrieval settings} For \emph{open} scientific claim verification, the system must retrieve candidate abstracts from a corpus of documents. In the \emph{abstract-provided} setting, candidate abstracts for each claim are given as input. We describe the retrieval settings for all datasets in \S \ref{sec:target_data}.

\paragraph{Supervision settings} We consider three supervision settings. In the \emph{zero-shot domain adaptation} setting, models may not train on any in-domain fact-checking data, though they may train on general-domain fact-checking data and other available scientific datasets. In the \emph{few-shot domain adaptation} setting, models may train on 45 claims from the target dataset. In the \emph{fully-supervised} setting, models may train on all claims from the target dataset. 

While most existing work on scientific fact-checking has focused on the fully-supervised setting, some recent work has examined the zero-shot setting. \citet{Lee2021TowardsFF} use language model perplexity as a measure of claim veracity. \citet{Wright2022GeneratingSC} generate claims based on citation sentences, and verify each generated claim against the abstracts mentioned in the claim's source citation. Given the high potential impact of fact verification systems for specialized domains, combined with the substantial cost of creating these datasets, we believe that the development of techniques for zero / few-shot domain adaptation represents an important area for continued research.

\subsection{Scientific claim verification datasets}

Several scientific claim verification datasets have been released in the past few years. \covidFact \cite{Saakyan2021COVIDFactFE} and \healthVer \cite{Sarrouti2021EvidencebasedFO} verify COVID-19 claims against scientific literature. \pubhealth \cite{Kotonya2020ExplainableAF} verifies public health claims against news and web sources. \scifact \citep{Wadden2020FactOF} verifies claims made in citations in scientific papers. \climatefever \cite{Diggelmann2020CLIMATEFEVERAD} verifies claims about climate change against Wikipedia. In this work, our focus is verifying claims against scientific literature. We therefore perform experiments on the \covidFact, \healthVer, and \scifact datasets. Preprocessing details and summary statistics for these datasets are included in \S \ref{sec:target_data}.

\subsection{Models} \label{sec:model_bg}

Motivated in part by the \sciver shared task \cite{Wadden2021OverviewAI} and leaderboard, a number of models have been developed for \scifact (the focus of the shared task). The two strongest systems on the shared task were \vertserini \cite{Pradeep2021ScientificCV} and \pjoint \cite{Li2021APM}, which we adopt as baselines. More recently, \arsjoint \cite{Zhang2021AbstractRS} achieved performance competitive with these two systems.\footnote{Recent progress can be found on the \href{https://leaderboard.allenai.org/scifact}{SciFact leaderboard}.}

Given a claim $c$ and candidate abstract $a$, these models make predictions in two steps. First, they predict rationales $\Rhat(c, a) = \{\rhat_1(c, a), \dots, \rhat_n(c, a)\}$ likely to contain evidence. Then, they make a label prediction $\yhat(c, f_R(\Rhat(c, a))$ based on the claim and predicted rationales, where $f_R$ is a function which creates a representation of the predicted rationales. 

While existing models share this general approach, they use different functions $f_R$ to construct rationale representations. For \vertserini, rationale selection and label prediction are performed by two separate T5-3B models, and $f_R$ concatenates the text of the selected rationales. As a result, the label predictor may not have access to all context needed to make a correct label prediction. \pjoint and \arsjoint attempt to address this issue by encoding the claim and full abstract (truncating to 512 tokens), and using these representations as the basis for both rationale selection and label prediction. The function $f_R$ consists of self-attention layers over the (globally-contextualized) token representations of the predicted rationales. Thus, \pjoint and \arsjoint can incorporate abstract-level context into label decisions. However, the mechanism by which this occurs is more complex than for our proposed system and requires rationale supervision for all training instances.

\section{The \sysname model} \label{sec:models}

We propose the \sysname model for full-context claim verification. In \S \ref{sec:model_arch}, we describe our modeling approach. Rather than predicting rationales $\Rhat(c, a)$ followed by the overall fact-checking label $\yhat(c, f_R(\Rhat(c, a)))$, we predict $\yhat(c, a)$ directly based on an encoding of the entire claim and abstract, and enforce consistency of $\Rhat(c, a)$ with $\yhat(c, a)$ during decoding. A similar idea has been shown to be effective on sentiment analysis and propaganda detection with token-level rationales \cite{Pruthi2020WeaklyAS}. In \S \ref{sec:fewshot_training}, we explain how our approach facilitates few-shot domain adaptation using weakly-labeled scientific documents.

\subsection{Full-context claim verification} \label{sec:model_arch}

\paragraph{Long-document encoding}

Given a claim $c$ and candidate abstract $a$ consisting of title $t$ and sentences $s_1, \dots, s_n$, we concatenate the inputs separated by$\tsl$tokens. The$\tsl$token following each sentence $s_i$ is notated as$\tslsub{i}$:
\begin{equation*}
  \ts c \tsl t \tsl s_1 \tslsub{1} \dots s_n \tslsub{n}
\end{equation*}
The model input sometimes exceeds the 512-token limit common to transformer-based language models like \bert \citep{Devlin2019BERTPO} and \roberta \citep{Liu2019RoBERTaAR}; see Table \ref{tbl:dataset_summary} for details on how frequently this occurs. Therefore, we use the \longformer model \cite{Beltagy2020LongformerTL} as our encoder. We assign global attention to the $\ts$ token, as well as all tokens in $c$ and all $\tsl$ tokens.

\paragraph{Multitask rationale selection and label prediction}

Given the full-context \longformer encoding, we predict whether sentence $s_i$ is a rationale via a binary classification head, consisting of two feedforward layers followed by a two-way softmax, on top of the globally-contextualized token $\tslsub{i}$. 

Similarly, we predict the overall fact-checking label $\yhat(c, a)$ by adding a three-way classification head over the encoding of the $\ts$ token. Since the $\ts$ token is trained with global attention, the model makes predictions based on a representation of the entire claim and abstract.

During training, we compute the cross-entropy losses for the label and rationale predictions, and train to minimize the multitask loss:
\begin{equation}  \label{eqn:loss}
  L  =  L_{\textrm{label}} + \lamRationale L_{\textrm{rationale}}
\end{equation}
where $\lamRationale$ is tuned on the dev set.

At inference time, we first predict $\yhat(c, a)$ to be the label with the highest softmax score. If the predicted label is \nei, we predict no rationales. If the predicted label is either \supports or \refutes, then we predict rationales as all sentences with an assigned softmax score of greater than 0.5. If no sentences have a rationale softmax over 0.5, then we predict the highest-scoring sentence as the sole rationale. In \S \ref{sec:ablations}, we show that this ability to condition the rationale predictions on the label prediction (as opposed to conditioning the label on the predicted rationales) leads to substantial improvement in the zero-shot domain adaptation setting.

\paragraph{Candidate abstract retrieval}

For datasets that require retrieval of candidate abstracts, we rely on the \vertserini \cite{Pradeep2021ScientificCV} retrieval system, which achieved state-of-the-art performance on the \sciver shared task (\sciver used the \scifact dataset for evaluation). This model first retrieves abstracts using BM25 \cite{Robertson2009ThePR}, then refines the predictions using a neural re-ranker based on \citet{Nogueira2020DocumentRW}, which is trained on the MS MARCO passage dataset \cite{Campos2016MSMA}.

\subsection{Training for domain adaptation} \label{sec:fewshot_training}

Three types of data are available to train scientific claim verification systems. (1) In-domain fact-checking annotations are the ``gold standard'', but they are expensive to create and require expert annotators. (2) General-domain fact-checking datasets like \fever \cite{Thorne2018FEVERAL} are abundantly available, but generalize poorly to scientific claims (see \S \ref{sec:results}). (3) Scientific documents -- either unlabeled or labeled for different tasks -- are abundant, and high precision heuristics (described in \S \ref{sec:prelim_finetuning}) can be used to generate document-level fact-checking labels $y(c, a)$ for these data. 

We train \sysname as follows: we first pretrain on a combination of general-domain fact-checking annotations, combined with weakly-labeled in-domain data.\footnote{
% ``Pretraining'' is a slight abuse of terminology. 
We use ``pretraining'' as shorthand for ``training on the target task with out-of-domain and/or weakly-supervised labels.''} Then, we finetune on the target scientific fact-checking dataset. The multitask architecture of \sysname is well-suited to this strategy, since the model can be trained on data with or without rationale annotations. When no rationales are available, we set $\lamRationale = 0$ in the loss function and train as usual. By contrast, training an ``extract-then-label'' model on weakly-supervised data requires creating rationale annotations $R(c, a)$, which is quite noisy (see \S \ref{sec:prelim_finetuning}).

\section{Datasets} \label{sec:datasets}

\begin{table*}[t]
  \footnotesize
  \setlength{\tabcolsep}{0.3em}
  \centering

  \begin{tabular}{{p{0.18\linewidth}p{0.08\linewidth}p{0.13\linewidth}R{0.04\linewidth}R{0.042\linewidth} p{0.10\linewidth}p{0.1\linewidth}R{0.06\linewidth}R{0.06\linewidth}R{0.07\linewidth}}}

  \toprule
  \tworows{Dataset} & \tworows{Domain} & \tworows{Claim source} & \tworows{Open} & \tworows{Has                                                \\ \nei} & \tworows{Claim  \\ complexity} & \tworows{Negation \\ method} &  \tworows{Train \\ claims} & \tworows{Eval \\ claims} & \tworows{> 512 \\ tokens}  \\
  \\
  \midrule
  \healthVer        & COVID            & TREC-COVID             & \xmark         & \cmark       & Complex & Natural   & 1,622   & 230 & 14.9\% \\
  \covidFact        & COVID            & Reddit                 & \xmark         & \xmark       & Complex & Automatic & 903     & 313 & 12.4\% \\
  \scifact          & Biomed           & Citations              & \cmark         & \cmark       & Atomic  & Human     & 1,109   & 300 & 27.4\% \\
  \midrule
  \fever            & Wiki             & Wikipedia              & -              & \cmark       & Atomic  & Human     & 130,644 & -   & 33.2\% \\
  \pqa              & Biomed           & Paper titles           & -              & \cmark       & Complex & Automatic & 58,370  & -   & 12.1\% \\
  \ei               & Biomed           & ICO prompts            & -              & \cmark       & Atomic  & Automatic & 7,395   & -   & 42.7\% \\
  \bottomrule
\end{tabular}

  \caption{Summary of datasets used in experiments. The top group of datasets are scientific claim verification datasets, and the bottom group are for pretraining. Datasets with a \cmark for ``Open'' require that candidate abstracts be retrieved from a corpus; those with a \xmark provide candidate abstracts as input. Datasets with a \cmark for  ``Has NEI'' require three-way (\supports\ / \refutes\ / \nei) label prediction, while those with an \xmark are (\supports\ / \refutes) only. The ``> 512 tokens'' column indicates the percentage of claim / abstract contexts that exceed 512 tokens.}

  \label{tbl:dataset_summary}

\end{table*}

\subsection{Scientific claim verification datasets} \label{sec:target_data}

We experiment with three scientific claim verification datasets. Table \ref{tbl:dataset_summary} provides a summary of important dataset characteristics. Preprocessing steps and additional statistics can be found in Appendix \ref{appx:data}.  \healthVer and \covidFact were originally released in an NLI format, pairing claims with (out-of-context) evidentiary sentences. We convert to our task format by identifying the abstracts in the CORD-19 corpus \cite{Wang2020CORD19TC} containing these sentences.

We use the following terminology: an \emph{atomic} claim makes an assertion about a single property of a single entity, while a \emph{complex} claim may make assertions about multiple properties or entities.

\paragraph{\scifact} Claims in \scifact \cite{Wadden2020FactOF} were created by re-writing citation sentences occurring in biomedical literature into atomic claims, which were verified against the abstracts of the cited documents. \refuted claims were created by manually negating the original claims. Abstracts that were cited but which annotators judged not to contain evidence were labeled \nei. \scifact requires retrieval of candidate abstracts.

\paragraph{\healthVer} \citep{Sarrouti2021EvidencebasedFO} consists of COVID-related claims obtained by extracting snippets from articles retrieved to answer questions from TREC-COVID \citep{Voorhees2020TRECCOVIDCA}, verified against abstracts from the CORD-19 corpus \cite{Wang2020CORD19TC}. Claims in \healthVer may be complex. \refuted claims occur naturally in the article snippets. \healthVer provides candidate abstracts for  each claim, but some of these candidates do not contain sufficient information to justify a \supports\ / \refutes verdict and are labeled \nei.

\paragraph{\covidFact} \cite{Saakyan2021COVIDFactFE} collects claims about COVID-19 scraped from a COVID-19 subreddit, and verifies them against linked scientific papers, as well as documents retrieved via Google search. Claims in \covidFact may be complex, and candidate abstracts for each claim are provided. All candidates either \support or \refute the claim. Claim negations were created automatically by replacing salient words in the original claims, and as a result the labels $y(c, a)$ are somewhat noisy (see Appendix \ref{appx:data}). 

\subsection{Pretraining datasets} \label{sec:prelim_finetuning}

We briefly describe our pretraining datasets and the weak supervision heuristics used to construct them. Detailed descriptions of these heuristics can be found in Appendix \ref{appx:data_processing}.

\paragraph{\fever} \cite{Thorne2018FEVERAL} consists of claims created by re-writing Wikipedia sentences into atomic claims, verified against Wikipedia articles.

\paragraph{\ei} \cite{Lehman2019InferringWM, DeYoung2020EvidenceI2} was released to facilitate understanding of clinical trial reports, which examine the effect of an \emph{intervention} on an \emph{outcome}, relative to a \emph{comparator} (``ICO'' elements). The dataset contains ICO \emph{prompts} paired with (1) labels indicating whether the outcome \emph{increased} or \emph{decreased} due to the intervention, and (2) rationales justifying each label. We use rule-based heuristics to convert these prompts into claims -- for instance ``[intervention] increases [outcome] relative to [comparator]''.

\paragraph{\pqa} \cite{Jin2019PubMedQAAD} was released to facilitate question-answering over biomedical research abstracts. We use the \pqaa subset, which is a large collection of abstracts with ``claim-like'' titles -- for instance, ``Vitamin B6 supplementation increases immune responses in critically ill patients.'' We treat the paper titles as claims and the matching abstracts as the evidence sources. 

To train models requiring rationale supervision, we create weakly-supervised rationales by selecting the sentences with highest similarity to the claim as measured by cosine similarity of Sentence-BERT embeddings \cite{Reimers2019SentenceBERTSE}. These annotations are not used to train \sysname. To estimate the precision of our rationale labeling heuristic, we predict rationales in the same fashion for our supervised datasets and compute the Precision@1 with which this method identifies gold rationales. The scores are relatively low: 49.4, 48.8, and 43.4 for \scifact, \covidFact, and \healthVer respectively. 

\section{Experimental setup} \label{sec:experimental_setup}

We describe our model training procedure, the systems against we compare \sysname, and our ablation experiments.

\subsection{Model training} \label{sec:model_training} 

Our complete training procedure consists of pretraining on the three datasets from \S \ref{sec:prelim_finetuning}, followed by finetuning on one of the target datasets from \S \ref{sec:target_data}. We conduct experiments with three different levels of supervision. For \emph{zero-shot} experiments, we perform pretraining only. For \emph{few-shot} experiments, we pretrain followed by finetuning on 45 target examples. For \emph{fully-supervised} experiments, we pretrain and then train on all target data.

Following \citet{Li2021APM}, we found that negative sampling was important to achieve good precision on \scifact, which requires document retrieval. We train with 20 negative samples per claim and retrieve 10 abstracts per claim at inference time. Appendix \ref{appx:negative_sampling} shows results without negative sampling. For the other datasets, no negative sampling was used.  Additional details including batch sizes, learning rates, number of epochs, etc. can be found in Appendix \ref{appx:model}.

During model development, we experimented with training on all three target datasets combined before predicting on each one, but found that this did not improve performance; see Appendix \ref{appx:generalization}. 

\begin{table*}[t]
  \footnotesize
  \setlength{\tabcolsep}{0.3em}
  \centering

  \begin{tabular}{
    p{0.07\linewidth}
    p{0.15\linewidth}
    P{0.026\linewidth}P{0.026\linewidth}P{0.033\linewidth}  P{0.026\linewidth}P{0.026\linewidth}P{0.033\linewidth} P{0.004\linewidth}
    P{0.026\linewidth}P{0.026\linewidth}P{0.033\linewidth}  P{0.026\linewidth}P{0.026\linewidth}P{0.033\linewidth} P{0.004\linewidth}
    P{0.026\linewidth}P{0.026\linewidth}P{0.033\linewidth}  P{0.026\linewidth}P{0.026\linewidth}P{0.033\linewidth} P{0.004\linewidth}
    }
    \toprule
                          & {}             & \multicolumn{6}{c}{\textbf{\healthVer}}                                       &  & \multicolumn{6}{c}{\textbf{\covidFact}}                                       &  & \multicolumn{6}{c}{\textbf{\scifact}}                                        & \\
                          & {}             & \multicolumn{3}{c}{Abstract}          & \multicolumn{3}{c}{Sentence}          &  & \multicolumn{3}{c}{Abstract}          & \multicolumn{3}{c}{Sentence}          &  & \multicolumn{3}{c}{Abstract}          & \multicolumn{3}{c}{Sentence}           \\
    \cmidrule(lr){3-5} \cmidrule(lr){6-8} \cmidrule(lr){10-12} \cmidrule(lr){13-15} \cmidrule(lr){17-19} \cmidrule(lr){20-22}
    \textbf{Setting}      & \textbf{Model} & \sz{P}    & \sz{R}    & F1            & \sz{P}    & \sz{R}    & F1            &  & \sz{P}    & \sz{R}    & F1            & \sz{P}    & \sz{R}    & F1            &  & \sz{P}    & \sz{R}    & F1            & \sz{P}    & \sz{R}    & F1             \\
    \midrule
    \multirow{2}{*}{Zero} & \pjoint        & \sz{72.3} & \sz{14.4} & 24.0          & \sz{22.9} & \sz{2.7}  & 4.9           &  & \sz{51.3} & \sz{37.9} & 43.6          & \sz{31.5} & \sz{16.0} & 21.3          &  & \sz{52.9} & \sz{32.4} & 40.2          & \sz{36.4} & \sz{14.9} & 21.1           \\
    \arrayrulecolor{black!30}\cmidrule(lr){2-22} \arrayrulecolor{black!100}
                          & \sysname       & \sz{60.6} & \sz{20.5} & \textbf{30.7} & \sz{25.0} & \sz{4.6}  & \textbf{7.8}  &  & \sz{48.8} & \sz{45.7} & \textbf{47.2} & \sz{32.7} & \sz{18.5} & \textbf{23.6} &  & \sz{49.0} & \sz{44.6} & \textbf{46.7} & \sz{39.0} & \sz{21.6} & \textbf{27.8}  \\
    \midrule
    \multirow{2}{*}{Few}  & \pjoint        & \sz{62.7} & \sz{41.6} & 50.0          & \sz{46.0} & \sz{29.3} & \textbf{35.8} &  & \sz{73.3} & \sz{60.6} & 66.3          & \sz{44.3} & \sz{30.6} & 36.2          &  & \sz{44.4} & \sz{51.4} & 47.6          & \sz{33.0} & \sz{35.1} & 34.0           \\
    \arrayrulecolor{black!30}\cmidrule(lr){2-22} \arrayrulecolor{black!100}
                          & \sysname       & \sz{63.6} & \sz{47.9} & \textbf{54.7} & \sz{41.9} & \sz{31.0} & 35.7          &  & \sz{71.3} & \sz{68.1} & \textbf{69.7} & \sz{39.5} & \sz{35.4} & \textbf{37.4} &  & \sz{76.4} & \sz{54.1} & \textbf{63.3} & \sz{51.7} & \sz{40.3} & \textbf{45.3}  \\
    \midrule
    \multirow{3}{*}{Full} & \vertserini    & \sz{71.3} & \sz{74.0} & 72.6          & \sz{65.6} & \sz{61.2} & 63.3          &  & \sz{76.6} & \sz{52.7} & 62.4          & \sz{44.8} & \sz{27.2} & 33.9          &  & \sz{64.0} & \sz{73.0} & 68.2          & \sz{60.6} & \sz{66.5} & 63.4           \\
                          & \pjoint        & \sz{75.0} & \sz{68.3} & 71.5          & \sz{69.9} & \sz{60.6} & 64.9          &  & \sz{71.5} & \sz{68.1} & 69.8          & \sz{41.4} & \sz{40.3} & 40.8          &  & \sz{75.8} & \sz{63.5} & 69.1          & \sz{68.9} & \sz{54.6} & 60.9           \\
    \arrayrulecolor{black!30}\cmidrule(lr){2-22} \arrayrulecolor{black!100}
                          & \sysname       & \sz{78.9} & \sz{76.3} & \textbf{77.6} & \sz{71.4} & \sz{67.0} & \textbf{69.1} &  & \sz{77.3} & \sz{77.3} & \textbf{77.3} & \sz{41.5} & \sz{46.1} & \textbf{43.7} &  & \sz{73.8} & \sz{71.2} & \textbf{72.5} & \sz{67.4} & \sz{67.0} & \textbf{67.2}  \\
    \bottomrule
\end{tabular}

  \caption{Performance of \sysname and baselines. In the fully-supervised setting, we compare to \pjoint and \vertserini, which exhibit comparable performance. In the zero and few-shot settings, we compare to \pjoint only due to the high cost of pretraining \vertserini.  We report performance using abstract-level and sentence-level evaluation as defined in \S \ref{sec:task_review}.}

  \label{tbl:main_results}

\end{table*}

% Include all ablations

\begin{table}[t!]
    \setlength{\tabcolsep}{0.3em}
    \footnotesize
    \centering

    \begin{subtable}[h]{\linewidth}
        \begin{tabular}{
                L{2em}
                L{4.85em}
                        *{3}{R{4.9em}}
        }
        \toprule
        \multicolumn{2}{r}{\textbf{Pretraining}} & \healthVer                    & \covidFact                    & \scifact                      \\
        \midrule
        \multirow{2}{*}{Zero} & \feverSci        & \textbf{30.7} / \textbf{7.8}  & \textbf{47.2} / \textbf{23.6} & \textbf{46.7} / \textbf{27.8} \\
                              & \fever           & 1.3 / 0.7                     & 25.2 / 11.2                   & 23.9 / 11.8                   \\
        \cmidrule(lr){1-5}
        \multirow{3}{*}{Few}  & \feverSci        & \textbf{54.7} / \textbf{35.7} & 69.7 / 37.4                   & \textbf{63.3} / \textbf{45.3} \\
                              & \fever           & 53.4 / 31.9                   & \textbf{74.4} / \textbf{42.1} & 54.5 / 39.0                   \\
                              & \targetOnly      & 39.4 / 27.0                   & 67.8 / 22.6                   & 24.2 / 10.8                   \\
        \cmidrule(lr){1-5}
        \multirow{3}{*}{Full}  & \feverSci        & \textbf{77.6} / 69.1          & 77.3 / \textbf{43.7}          & \textbf{72.5} / \textbf{67.2} \\
                              & \fever           & 77.1 / \textbf{70.3}          & \textbf{77.4} / 43.3          & 67.9 / 61.7                   \\
                              & \targetOnly      & 74.5 / 69.7                   & 69.7 / 36.6                   & 63.3 / 58.4                   \\
        \bottomrule
\end{tabular}

        \subcaption{Effect of pretraining data. In-domain pretraining is very effective in the zero- and few-shot settings. In the zero-shot setting, ``\targetOnly'' metrics are not shown since this would correspond to no training at all.}
        \label{tbl:ablations_pretrain}
        \vspace{1em}
    \end{subtable}

    \begin{subtable}[h]{\linewidth}
        \begin{tabular}{
                L{2em}
                L{4.85em}
                        *{3}{R{4.9em}}
        }
        \toprule
                              & \textbf{Encoder}
                              & \healthVer       & \covidFact                    & \scifact                                                      \\
        \midrule
        \multirow{2}{*}{Zero} & Longformer       & 30.7 / 7.8                    & 47.2 / 23.6                   & \textbf{46.7} / \textbf{27.8} \\
                              & RoBERTa          & \textbf{34.2} / \textbf{9.2}  & \textbf{48.3} / \textbf{26.2} & 45.2 / 25.9                   \\
        \cmidrule(lr){1-5}
        \multirow{2}{*}{Few}  & Longformer       & \textbf{54.7} / 35.7          & 69.7 / 37.4                   & \textbf{63.3} / \textbf{45.3} \\
                              & RoBERTa          & 51.2 / \textbf{36.9}          & \textbf{72.1} / \textbf{41.0} & 50.5 / 34.0                   \\
        \cmidrule(lr){1-5}
        \multirow{2}{*}{Full} & Longformer       & 77.6 / 69.1                   & 77.3 / \textbf{43.7}          & \textbf{72.5} / \textbf{67.2} \\
                              & RoBERTa          & \textbf{78.8} / \textbf{72.7} & \textbf{78.2} / 43.4          & 67.6 / 62.3                   \\
        \bottomrule
\end{tabular}

        \subcaption{Effect of base encoder. \longformer improves performance on \scifact, which has the largest fraction of instances exceeding the \roberta token limit.}
        \label{tbl:ablations_encoder}
        \vspace{1em}
    \end{subtable}

    \begin{subtable}[h]{\linewidth}
        \begin{tabular}{
                L{2em}
                L{4.85em}
                        *{3}{R{4.9em}}
        }
        \toprule
                              & \textbf{Approach}
                              & \healthVer        & \covidFact                    & \scifact                                                      \\
        \midrule
        \multirow{3}{*}{Zero} & Multitask         & \textbf{30.7} / \textbf{7.8}  & \textbf{47.2} / \textbf{23.6} & \textbf{46.7} / \textbf{27.8} \\
                              & Pipe              & 3.2 / 0.9                     & 19.0 / 10.5                   & 22.5 / 12.8                   \\
                              & MT / PI           & 4.5 / 1.8                     & 26.7 / 13.5                   & 28.3 / 17.7                   \\
        \cmidrule(lr){1-5}
        \multirow{3}{*}{Few}  & Multitask         & \textbf{54.7} / \textbf{35.7} & \textbf{69.7} / 37.4          & \textbf{63.3} / \textbf{45.3} \\
                              & Pipe              & 52.8 / 29.5                   & 68.3 / \textbf{38.2}          & 53.0 / 39.9                   \\
                              & MT / PI           & 46.7 / 32.3                   & 59.3 / 34.1                   & 56.2 / 41.1                   \\
        \cmidrule(lr){1-5}
        \multirow{3}{*}{Full} & Multitask         & 77.6 / 69.1                   & 77.3 / 43.7                   & \textbf{72.5} / \textbf{67.2} \\
                              & Pipe              & \textbf{78.4} / \textbf{69.2} & \textbf{77.6} / \textbf{47.7} & 70.9 / 66.2                   \\
                              & MT / PI           & 70.6 / 64.3                   & 73.3 / 44.0                   & 60.3 / 57.0                   \\
        \bottomrule
\end{tabular}

        \subcaption{Effect of model architecture. The Multitask approach performs best in the zero- and few-shot settings. We examine the fully-supervised setting in detail in \S \ref{sec:analysis_pipeline}.}
        \label{tbl:ablations_architecture}
    \end{subtable}

    \caption{
        Ablations examining the effects of pretraining data, base encoder, and modeling approach. Entries are formatted ``\{Abstract-level F1\} / \{Sentence-level F1\}''. 
    }
    \label{tbl:ablations}
\end{table}

\subsection{Baseline systems} \label{sec:baselines}

We use \pjoint and \vertserini as baselines. \vertserini is the largest model, with 5.6B parameters. \sysname and \pjoint are comparably-sized, with 440M and 360M parameters, respectively.

In the fully-supervised setting, we compare against both baselines. 
For prediction on \scifact, we use publicly available model checkpoints as-is. 
For training on \healthVer and \covidFact, we use the code provided by the authors, starting from the available checkpoints trained on \scifact. Model hyperparameters (learning rate, batch size, epoch number, etc.) for all systems including \sysname were tuned based solely on \scifact and not adjusted further. Additional details can be found in Appendix \ref{appx:baselines}.
% For a fair comparison with the baselines (which were designed for \scifact), we performed model development for \sysname on \scifact as well, and did not modify the training procedure for the other two datasets. 

Evaluation in the few-shot and zero-shot settings requires pretraining and finetuning as described in \S \ref{sec:model_training}. Due to the expense of pretraining T5-3B, we do not perform these experiments for \vertserini, and compare only against \pjoint (which shows comparable performance in the fully-supervised setting). We pretrain \pjoint on the data described in \S \ref{sec:prelim_finetuning}.
% , using rationales selected by Sentence-BERT when pretraining on \pqa. 

\subsection{Ablations} \label{sec:ablations_setup}

Since \pjoint and \vertserini differ from \sysname along a number of important dimensions (e.g. model architecture, number of parameters, and base encoder), we conduct ablations to characterize the performance contributions of three key components of \sysname.

\paragraph{Pretraining data} We compare the results of three different pretraining strategies. For \feverSci, we pretrain on all available data as described in \S \ref{sec:model_training}. For \fever, we pretrain on \fever only. For \emph{\targetOnly}, we perform no pretraining.

\paragraph{Base encoder} We compare the performance achieved using LongFormer as the encoder for \sysname, compared to the results when we swap in \roberta but keep other settings identical. We use \longformer-large and \roberta-large.

\paragraph{Modeling approach} We compare three modeling approaches: (1) The \emph{Multitask} approach is the method used by \sysname as described in \S \ref{sec:model_arch}. (2) The \emph{Pipeline} approach consists of two separate \longformer modules. The first selects rationales as described in \S \ref{sec:model_arch}, but with $L_{\textrm{label}}$ removed from Eq.~\ref{eqn:loss}, and the second module predicts a label given the text of the rationales selected by the first module as input. When pretraining on \pqa, we train on the rationales chosen by Sentence-BERT as described in \S \ref{sec:prelim_finetuning}. (3) The \emph{Multitask train / Pipeline inference} (MT / PI) approach takes the model trained using the Multitask approach, and performs inference using the Pipeline approach. Specifically, MT / PI is trained to make label predictions based on full abstracts, but must make test-time label predictions based on predicted rationales only. By contrast, the Pipeline model makes label predictions based on gold and predicted rationales at train and test time, respectively.

\section{Experimental results} \label{sec:experiments}

We compare \sysname performance relative to our baseline systems, and present ablation results.

\subsection{Main Results} \label{sec:results}

Table \ref{tbl:main_results} compares the performance of \sysname against \pjoint and \vertserini. A few trends are apparent. First, \sysname outperforms the baselines on all datasets, with relative improvements --- averaged over the three datasets and two evaluation methods --- of 26\%, 14\%, and 11\% in the zero-shot, few-shot, and fully-supervised settings respectively. We examine possible causes of this improvement in \S \ref{sec:ablations}. Second, while all models score within roughly six points of each other on \healthVer and \scifact, variability is much greater on \covidFact. We suspect that this is due to the automatically-generated nature of \covidFact negations. Third, we observe that \healthVer appears to be the most challenging dataset of the three. Few-shot abstract-level F1 scores for \covidFact and \scifact are generally within 10 F1 of their fully-supervised values, while the gap is roughly 20 F1 for \healthVer. This may be due to the high complexity of \healthVer claims. 

\subsection{Ablations} \label{sec:ablations}

The results of all ablations are shown in Table \ref{tbl:ablations}. We report abstract and sentence-level F1 scores in the main text; full results can be found in Table \ref{tbl:ablations_full} in Appendix \ref{appx:results}.

\paragraph{In-domain pretraining substantially improves zero / few-shot performance}

In Table \ref{tbl:ablations_pretrain}, we compare the performance of models pretrained on \feverSci, \fever, and \targetOnly. In the zero-shot setting, removing scientific data during pretraining results in a relative performance decrease of 65\%, averaged over the three datasets and two evaluation methods. The decrease is driven primarily by very low recall (see Table \ref{tbl:ablations_pretrain_full}).

In the few-shot setting, \fever pretraining scores within 4\% of \feverSci, while \targetOnly results in a 39\% decrease relative to \feverSci. This suggests that training on a handful of target examples is sufficient to recalibrate a model trained for a different domain, but not to learn the task from scratch. In the fully-supervised setting, \fever pretraining is only slightly worse than \feverSci, while \targetOnly lags by roughly 9\%. Overall, the results indicate that pretraining always helps, and pretraining on weakly-labeled in-domain data helps especially when target data are scarce.

\paragraph{\longformer improves performance on datasets with long documents}

Table \ref{tbl:ablations_encoder} compares the performance of \sysname when \longformer and \roberta are used as the base encoder. Using \longformer consistently helps on \scifact, but does not help on the other two datasets. This is unsurprising, since 27\% of \scifact instances exceed the \roberta token limit, compared to less than 15\% for the other two datasets (Table \ref{tbl:dataset_summary}). 

\paragraph{Multitask modeling improves zero / few-shot performance}

Results comparing our three different modeling approaches are shown in Table \ref{tbl:ablations_architecture}. In the zero-shot setting, we find that Multitask performs best, with both MT / PI and Pipeline exhibiting performance drops greater than 50\%. The Multitask approach of predicting rationales conditioned on the predicted label leads to improved recall (see Table \ref{tbl:ablations_architecture_full}). Similarly, in the few-shot setting, both Pipeline and MT / PI perform roughly 10\% worse than Multitask. Collectively, the results indicate that Multitask makes the best use of the available data when target annotations are limited.

We also find that MT / PI outperforms Pipeline in the zero-shot setting. This supports our intuition from \S \ref{sec:fewshot_training} that, while training on weakly-supervised \emph{document-level} labels improves zero-shot performance, training on weakly-supervised \emph{sentence-level} rationales (as for Pipeline) leads to worse performance than not training on these rationales (as for MT / PI).  

In the fully-supervised setting, Multitask performs best on \scifact, while Pipeline slightly outperforms Multitask on \healthVer and \covidFact. MT / PI performs substantially worse than the other approaches on all datasets. We investigate these findings further in \S \ref{sec:analysis_pipeline}; our results indicate that Pipeline may, in effect, be trained to make predictions based on insufficient evidence.

\section{Analysis} \label{sec:analysis}

\subsection{Fully-supervised Pipeline performance} \label{sec:analysis_pipeline}

\begin{table}[t]
    \setlength{\tabcolsep}{0.3em}
    \footnotesize
    \centering

    \begin{tabular}{
        L{4.5em}
        *{3}{P{1.3em}}
        P{0.5em}
        *{3}{P{1.3em}}
        P{0.5em}
        P{3em}
    }
    \toprule
    {}                & \multicolumn{3}{c}{\tworowsc{\textbf{Self-} \\ \textbf{contained}}} &  & \multicolumn{3}{c}{\tworowsc{\textbf{Context-} \\ \textbf{dependent}}} &  &                      \\
    \\
    \cmidrule(lr){2-4}  \cmidrule(lr){6-8}
    \textbf{Approach} & \sz{P}    & \sz{R}    & F1                                          &  & \sz{P}     & \sz{R}    & F1                                            &  & $\mathbf{\% \Delta}$ \\
    \midrule
    Multitask         & \sz{86.1} & \sz{82.9} & 84.5                                        &  & \sz{90.3}  & \sz{60.9} & 72.7                                          &  & -14.0\%              \\
    Pipeline          & \sz{92.4} & \sz{89.0} & 90.7                                        &  & \sz{82.4}  & \sz{60.9} & 70.0                                          &  & -22.8\%              \\
    MT / PI           & \sz{91.8} & \sz{54.9} & 68.7                                        &  & \sz{100.0} & \sz{13.0} & 23.1                                          &  & -66.4\%              \\
    \cmidrule(lr){1-10}
    Count             & \multicolumn{3}{l}{82}                                              &  & \multicolumn{3}{l}{46}                                                 &  &                      \\
    \bottomrule
\end{tabular}

    \caption{
        Performance of the Multitask, Pipeline, and MT / PI modeling approaches on \scifact instances with rationales that are self-contained (can be interpreted in isolation) or context-dependent (must be interpreted in the context of the abstract). Evaluation is performed in the abstract-provided setting. We report abstract-level metrics; sentence-level results are similar. The $\% \Delta$ indicates the drop in F1 score on context-dependent instances relative to self-contained instances. Multitask suffers the smallest performance loss, while MT / PI suffers the largest.
    }
    \label{tbl:skills_context}
\end{table}

In \S \ref{sec:ablations}, we found that the Pipeline approach (but not the MT / PI approach) performed on par with the Multitask approach in the fully-supervised setting. To understand this finding, we collected detailed annotations for 128 claim / evidence instances from the \scifact test set. For each instance, an annotator indicated whether the annotated rationales were ``self-contained'' --- i.e. sufficient to justify the fact-checking label when taken in isolation, or  ``context-dependent'' --- i.e. only sufficient when taken in the context of the abstract. Figure \ref{fig:teaser} and Table \ref{tbl:reasoning_categories} provide examples; see \citet{Choi2021DecontextualizationMS} for a detailed discussion of different forms of context-dependence.\footnote{Unlike \citet{Choi2021DecontextualizationMS}, we do not include the presence of acronyms as ``context-dependent,'' since an acronym can be matched with its expansion based on surface-level textual features. See Appendix~\ref{appx:acronyms} for further analysis of acronyms.}

Table \ref{tbl:skills_context} compares the performance of the three modeling approaches on instances with self-contained vs.~context-dependent evidence. We find that all approaches have lower performance on context-dependent instances relative to self-contained instances, but the size of the performance drop varies widely. The Multitask approach performs 14.0\% worse on context-dependent instances, while the Pipeline approach performs 22.8\% worse. Most interestingly, MT / PI performs 66.4\% worse, driven predominantly by low recall. The MT / PI model frequently (and correctly) predicts that context-dependent rationales are not sufficient to justify a \supports\ / \refutes decision. These findings suggest that (1) the Multitask approach is, as expected, best at verifying claims with context-dependent evidence, and (2) the Pipeline approach has, in effect, over-fit to context-dependent rationales and learned to make predictions based on insufficient evidence.

\subsection{Performance upper bound}

To determine an ``upper bound'' on the achievable performance of scientific fact-checking models, we assign \numHumanAnnots claim-evidence pairs from \scifact for independent annotation by two different annotators. We estimate human-level performance by treating the first annotator's results as ``gold,'' and the second annotator's results as predictions. For comparison, we make predictions using \sysname and our two baseline models, with candidate abstracts provided as input. The results are shown in Table \ref{tbl:human_performance}. Existing systems already exceed human agreement for sentence-level evaluation, but not abstract-level, indicating that experts tend to agree on the overall relationship between claim and abstract, but may disagree about exactly which sentences contain the best evidence. This constitutes another reason not to rely solely on selected rationales when predicting a fact-checking label: the choice of rationales is itself somewhat subjective.

In addition, these results suggest that one key subtask of scientific claim verification --- namely, predicting whether an evidence-containing abstract \supports or \refutes a claim --- may be nearly ``solved'' in the setting where (1) the claims are atomic and (2) roughly 1,000 in-domain labeled claims are available for training.

\begin{table}[t]
  \footnotesize
  \setlength{\tabcolsep}{0.3em}
  \centering

  \begin{tabular}{
    p{0.30\linewidth}
    P{0.07\linewidth}P{0.07\linewidth}P{0.080\linewidth}P{0.0005\linewidth}
    P{0.07\linewidth}P{0.07\linewidth}P{0.080\linewidth}
    }
    \toprule
    {}          & \multicolumn{3}{c}{Abstract} &           & \multicolumn{3}{c}{Sentence}                                            \\
    \cmidrule(lr){2-4} \cmidrule(lr){6-8}
    {}          & \sz{P}                       & \sz{R}    & F1                           &  & \sz{P}    & \sz{R}    & F1            \\
    \midrule
    \vertserini & \sz{90.7}                    & \sz{74.3} & 81.7                         &  & \sz{79.6} & \sz{62.2} & 69.8          \\
    \pjoint     & \sz{87.2}                    & \sz{64.4} & 74.1                         &  & \sz{76.7} & \sz{55.1} & 64.1          \\
    \sysname    & \sz{87.4}                    & \sz{75.2} & 80.9                         &  & \sz{80.5} & \sz{70.3} & \textbf{75.0} \\
    \midrule
    Human       & \sz{94.8}                    & \sz{84.1} & \textbf{89.1}                &  & \sz{67.4} & \sz{67.4} & 67.4          \\
    \bottomrule
\end{tabular}

  \caption{Performance on \scifact in the ``abstract-provided'' setting. Models exceed human agreement as measured by sentence-level F1, but not abstract-level.}

  \label{tbl:human_performance}

\end{table}

\section{Related work}

Background on scientific claim verification is covered in \S \ref{sec:background}; we discuss other relevant work here. \citet{Nye2020UnderstandingCT} have previously observed that document-level context is often required to properly interpret scientific findings. 
% Relatedly, decontextualization has been studied more generally in \citet{Choi2021DecontextualizationMS}.

\citet{DeYoung2020EvidenceI2} use an ``extract-then-label'' pipeline for the original \ei task. Multitask label prediction and rationale selection was proposed by \citet{Pruthi2020WeaklyAS} and applied to sentiment analysis and propaganda detection. As in this work, the authors condition on the predicted label when predicting rationales.  Another alternative to supervised rationale selection is to treat rationales as latent variables \cite{Lei2016RationalizingNP,Paranjape2020AnIB}. 

Long-document encodings for fact verification have been explored by \citet{Stammbach2021EvidenceSA}, who use Big Bird \citep{Zaheer2020BigBT} for full-document evidence extraction from \fever. Domain adaptation for scientific text has been studied in a number of works, including \citet{Gururangan2020DontSP,Beltagy2019SciBERTAP,Lee2020BioBERTAP,Gu2022DomainSpecificLM}. In those works, the primary focus is on language model pretraining. Here, we focus on training on the target task using out-of-domain and weakly-labeled data.

\section{Conclusion}

This work points to a number of promising future directions for scientific claim verification. These include applying the approach presented here to develop scientific claim verification models for new scientific sub-domains or other specialized fields given a handful of labeled examples, and extending the task definition to verify claims against longer contexts (e.g. full scientific papers) or larger corpora. Our task formulation also offers an opportunity to study the effects of rationale decontextualization \citep{Choi2021DecontextualizationMS}, especially in cases where models may be making predictions based on insufficient evidence.

In presenting the \sysname system, we addressed two challenges associated with scientific claim verification: incorporating relevant information beyond rationale boundaries by modeling full-document context, and facilitating zero / few-shot domain adaptation through weak supervision enabled by a multitask modeling approach. 
Our experiments show that \sysname outperforms existing systems across several scientific claim verification datasets.
% \sysname outperformed existing systems across several science fact checking datasets, demonstrating the efficacy of our multitask approach and heuristic labeling methods. 
% kylel - commented out because somewhat overlaps first sentence
We hope that the task, data, and modeling resources presented in this paper will encourage further work and progress towards the broader goals of identifying and addressing scientific mis- and disinformation.

% through ablation studies and analysis.

% such as the further development of methods for few-shot domain adaptation, characterization of the performance of fact-checking models when verifying claims against realistic-sized corpora of millions of documents, and extending the approach developed here to contexts beyond scientific research abstracts, e.g., full scientific papers, news articles, or social media posts. Another promising alternative to the approach taken here would be to explicitly ``decontextualize'' evidence-containing rationales by filling in omitted context, and then make label predictions based on the decontextualized evidence. This approach could lead to modeling decisions that are justified by the evidence, while also being interpretable and easily explainable to users. 

% adding more tasks
% full document
% task formulation in which to study the impact of decontextualization for rationales
\section{Ethical considerations and broader impact}

One long-term goal of research on scientific claim verification is to build systems that can automatically identify mis- and dis-information, which we believe would be socially beneficial given the current prevalence of mis- and dis-information online.

In the shorter term, this work presents two potential risks. First, automated systems for scientific fact-checking are not mature enough to inform real-world medical decisions. We will include a disclaimer with released software to this effect. Second, bad actors could potentially use this work to develop disinformation generators trained to ``fool'' automated fact-checkers. While this risk cannot be ruled out, we believe that the benefits of publishing this work and making our models available to the community to facilitate further research outweigh the risks that this work will be misused by malicious actors.

\section*{Acknowledgments}

This research was supported by the ONR MURI N00014-18-1-2670, ONR N00014-18-1-2826, DARPA N66001-19-2-4031, NSF (IIS 1616112), NSF Convergence Accelerator Award \#2132318, an Allen Distinguished Investigator Award, and the Sloan fellowship. We thank the Semantic Scholar team at AI2, UW-NLP, and the H2lab at UW for helpful comments and feedback. Thanks to Arkadiy Saakyan and Tuhin Chakrabarty for help with \covidFact, to Mourad Sarrouti for help with \healthVer, and to Xiangci Li and Ronak Pradeep for help with \pjoint and \vertserini, respectively.

\bibliography{anthology,custom}
\bibliographystyle{acl_natbib}

% Appendices
\appendix
\section{Data processing and statistics} \label{appx:data}

\subsection{Data preprocessing} \label{appx:data_processing}

\paragraph{\scifact} We use \scifact in its original form, as it was released by the paper authors \cite{Wadden2020FactOF}.

\paragraph{\healthVer} The \healthVer \cite{Sarrouti2021EvidencebasedFO} data release available at \url{https://github.com/sarrouti/HealthVer} appears in NLI format, pairing claims with evidence-containing sentences; the documents from which the sentences were extracted are not provided. We match evidence-containing sentences to their abstracts in the CORD-19 corpus \cite{Wang2020CORD19TC} using a simple substring search, after normalizing for capitalization and whitespace differences. Evidence for which no match was found in the corpus is discarded. 

We then segment the abstracts into sentences. Any sentence in the abstract with a string overlap of $> 50 \%$ with the evidence provided in the original data is marked as a rationale. A small number of claims in \healthVer had both supporting and refuting evidence in the same abstract; we remove these claims as well to conform to our task definition. Modeling conflicting evidence is a promising direction for future work.

\paragraph{\covidFact} The \covidFact data available at \url{https://github.com/asaakyan/covidfact} is released in a similar format to \healthVer. Like \healthVer, we perform string search over CORD-19 to identify the abstracts containing evidence, and use the same procedure for assigning rationale labels to sentences from the abstract. \covidFact also includes evidence from sources scraped from the web that are not contained in CORD-19, such as news articles. These sources are not provided with the data release; we discard evidence from non-CORD-19 sources\footnote{Upon request, the paper authors did kindly provide us with scraped evidence documents. Unfortunately, we did not have time to re-run our experiments on these additional sources.}.
% TODO how much did we discard?

Refuted claims in \covidFact are generated automatically by replacing words in the original claim. Based on a manual inspection, we found this process to generate a truly refuted claim roughly a third of the time; in most other cases, it generated a claim that was either ungrammatical or for which the provided evidence was irrelevant. A few cases are provided in Table \ref{tbl:covid_fact}.

\begin{table*}[t] 
  \setlength{\tabcolsep}{2pt}
  \footnotesize
  \centering
  \begin{tabular}{L{0.3\linewidth} L{0.3\linewidth} L{0.33\linewidth}}
    \toprule
    \textbf{Original claim} &  \textbf{Automatic negation} & \textbf{Comment} \\
    \midrule
    Sars-cov-2 reactive t cells \dots are likely expanded by beta-coronaviruses & Sars-cov-2 reactive t cells \dots are not expanded by beta-coronaviruses & Successful negation \\
    \midrule
    Regn-cov2 antibody cocktail prevents and treats sars-cov-2 \dots & On-cov2 antibody cocktail prevents and treats sars-cov-2 infection \dots &  Ungrammatical; ``On-cov2'' isn't a scientific entity.  \\
    \midrule
    \dots immunity is maintained at 6 months following primary infection & \dots immunity is maintained at 6 weeks following primary infection & Not refuted; The original claim entails the negation. Immunity at 6 months implies immunity at 6 weeks. \\
    \bottomrule
  \end{tabular}
  \caption{Automatic negations from \covidFact. Some are successful, in the sense that the attempted negation contradicts the original claim. Others are either ungrammatical or are entailed by the original claim.}
  \label{tbl:covid_fact}
\end{table*}

\paragraph{FEVER}

We use the \fever dataset as-is.

\paragraph{\ei}

\begin{figure}[t]
  \centering
  \includegraphics[width=\columnwidth]{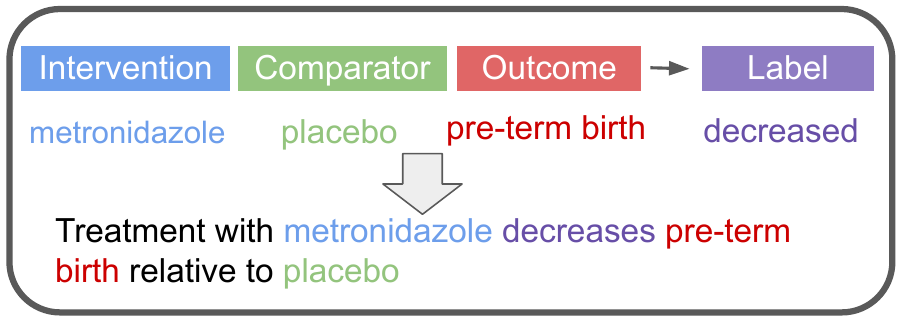}

  \caption{An example showing how an evidence inference prompt (top) can be converted into a claim (bottom) using templates. A refuted claim could be generated by substituting ``increases'' for ``decreases'' in the prompt text.}
  \label{fig:evidence_inference}
\end{figure}

The \ei dataset consists of ``ICO'' (intervention / comparator / outcome) prompts, paired with labels indicating whether the intervention leads to an increase, decrease, or no change in the outcome with respect to the comparator. The dataset is available at \url{https://evidence-inference.ebm-nlp.com/}. We use templates to convert these prompts to claims. See Figure \ref{fig:evidence_inference} for an example. Rationale annotations are provided for this dataset. Additional examples of templates are below; the full list will be included in the code release. Refuted claims are generated by swapping ``increase'' and ``decrease'' templates. 

\begin{itemize}[noitemsep]
    \item \textbf{Increase}: [intervention] raises [outcome] relative to [comparator]
    \item \textbf{No change}: [intervention] and [comparator] have very similar effects on [outcome]
    \item \textbf{Decrease}: [intervention] results in a decrease in [outcome], relative to [comparator]
\end{itemize}

\paragraph{\pqa} We use the PQA-A subset released at \url{https://pubmedqa.github.io/}, which is filtered for ``claim-like'' titles. We generate negations by identifying titles with the phrases ``does not'', ``do not'', ``are not'', ``is not''. ``Does not'' and ``do not'' are removed and the relevant verbs are modified to have the correct inflection; for instance ``smoking does not cause cancer'' is converted to ``smoking causes cancer''. Similarly, ``are not'' and ``is not'' are replaced by ``are'' and ``is''.

To generate rationales needed to train pipeline models on \pqa, we employ the following procedure. First, we encode the claim and all abstract sentences using the \texttt{all-MiniLM-L6-v2} model from the Sentence-Transformers package \url{https://www.sbert.net/}. Then, we rank abstract sentences by cosine similarity with the claim and label the top-$k$ sentences as rationales, where $k$ is randomly sampled from $\{1, 2, 3\}$ with a 4:2:1 frequency ratio (this matches the distribution of $k$ in \scifact). 

\subsection{Dataset statistics}

Table \ref{tbl:dataset_counts} provides counts showing the number of claim / evidence pairs with each label (\supports, \refutes, \nei), in each of our target datasets. Note that a given claim may be (and often is) paired with more than one abstract containing evidence. \healthVer is the largest dataset. \covidFact is the smallest, in part due to the aggressive evidence filtering described in \S \ref{appx:data_processing}.

\begin{table}[t]
  \footnotesize
  \setlength{\tabcolsep}{0.3em}
  \centering

  \begin{tabular}{{p{0.12\linewidth}p{0.25\linewidth}R{0.15\linewidth}R{0.15\linewidth}R{0.15\linewidth}}}
  \toprule
  Fold                   & Dataset    & \supports & \nei & \refutes \\
  \midrule
  \multirow{3}{*}{Train} & \scifact   & 508       & 485  & 265      \\
                         & \covidFact & 299       & -    & 641      \\
                         & \healthVer & 2384      & 2384 & 1464     \\
  \cmidrule(lr){1-5}
  \multirow{3}{*}{Eval}  & \scifact   & 113       & 127  & 109      \\
                         & \covidFact & 102       & -    & 215      \\
                         & \healthVer & 374       & 304  & 225      \\
  \bottomrule
\end{tabular}

  \caption{Evidence distribution by dataset.}

  \label{tbl:dataset_counts}

\end{table}

\subsection{Examples of context-dependent rationales} \label{appx:context_dependent}

Table \ref{tbl:reasoning_categories} provides an example of a context-dependent rationale (as defined in \S \ref{sec:analysis_pipeline}), as well as an example of a rationale with an undefined acronym. The latter occurs when an acronym appears in a rationale but its full expansion does not; an analysis of undefined acronyms is included in Appendix \ref{appx:acronyms}. The code and data release will contain full annotations indicating which of the 128 human-annotated examples described in \S \ref{sec:analysis_pipeline} are context-dependent, and which contain undefined acronyms.

\begin{table*}[t]
  % Commands for this table.
  \setlength{\tabcolsep}{2pt}
  \footnotesize
  \centering
  \begin{tabular}{L{0.13\linewidth} L{0.13\linewidth} L{0.71\linewidth}}
    \toprule
    \textbf{Category}                                                          & \multicolumn{2}{l}{\textbf{Example}}                                                                                                  \\
    \midrule
    \multirow{5}{*}{\shortstack[l]{\textbf{Context-} \\ \textbf{dependent}}} & \textbf{Claim:}       & \hl{Errors in peripheral IV drug administration} are most common during bolus administration                  \\
                                                                               & \textbf{Context:}     & \emph{OBJECTIVES: To determine the incidence of \hl{errors in the administration of intravenous drugs \dots}} \\
                                                                               & \textbf{Evidence:}    & \emph{\dots Most \hl{errors} occurred when giving bolus doses}                                                \\
                                                                               & \textbf{Explanation:} & The evidentiary sentence reporting the finding does not specify the type of \hl{error}.                       \\
    \midrule
    \multirow{4}{*}{\shortstack[l]{\textbf{Undefined} \\ \textbf{acronym}}}  & \textbf{Claim:}       & \hl{Hematopoietic stem cells} segregate their chromosomes randomly.                                           \\
                                                                               & \textbf{Context:}     & \emph{we tested these hypotheses in \hlc{lightblue}{hematopoietic stem cells (HSCs)}\dots}                    \\
                                                                               & \textbf{Evidence:}    & \emph{\dots indicated that all \hlc{lightgreen}{HSCs} segregate their chromosomes randomly.}                  \\
                                                                               & \textbf{Explanation:} & \hlc{lightgreen}{HSCs} is an acronym for \hlc{lightblue}{Hematopoietic stem cells}.                           \\
    \bottomrule
  \end{tabular}
  \caption{Examples from the \scifact dataset showcasing rationales that are context-dependent (top example), or include an undefined acronym (bottom example).}
  \label{tbl:reasoning_categories}

\end{table*}

\subsection{Annotators}

In \S \ref{sec:analysis}, we report an analysis based on annotations performed on the \scifact dataset. These annotations were performed by students and / or professional annotators associated with the authors' research institutions. Annotators were paid between \$15 and \$20 / hour.

\section{Modeling details} \label{appx:model}

\subsection{Implementation}

We implement \sysname using PyTorch Lightning (\url{https://www.pytorchlightning.ai/}), which relies on PyTorch (\url{https://pytorch.org/}). 

\subsection{Model training}

\paragraph{Pretraining} For pretraining, we train for 3 epochs on \fever, \ei, and \pqa, with the data randomly shuffled. We train on 4 negative samples (i.e. abstracts containing no evidence) per claim, which we find improves precision. We train on 8 NVIDIA RTX 6000 GPUs with a batch size of 1 / gpu (effective batch size of 8), using a learning rate of $1e-5$, using the PyTorch Lightning implementation of the AdamW optimizer with default settings. We initialize from a \longformer-large checkpoint pretrained on the \storc corpus \cite{Lo2020S2ORCTS}. 

\paragraph{Finetuning} For finetuning, we train for 20 epochs on the target dataset (\scifact, \healthVer, or \covidFact). For \scifact, we train on 20 negative samples / claim. To create ``hard'' negatives --- i.e. abstracts that have high lexical overlap with the claim --- we create a search index from 500K abstracts randomly selected from the biomedical subset of the \storc corpus. For each claim, we obtain negative abstracts by using the \vertserini retrieval system from \S \ref{sec:model_arch} to retrieve the top-1000 most-similar abstracts from this index, removing abstracts that are annotated as containing evidence, and randomly sampling 20 abstracts to be used as negatives during training.

Since \healthVer and \covidFact do not have a retrieval step, they do not require negative sampling, and we train on the original datasets as-is.

\paragraph{Retrieval} For \scifact, we performed dev set experiments retrieving 10, 20, or 50 abstracts / claim, and found that 10 was the best. We use that in our final experiments.

\subsection{Model hyperparameters}

No organized hyperparameter search was performed. We consulted with the authors of the \longformer paper for suggestions about good model parameters, and generally followed their suggestions.

The loss function in Section \ref{sec:model_arch} requires a weight $\lamRationale$. This is set to 15 for all final experiments. We informally experimented with values of 1, 5, and 15; no organized hyperparameter search was performed.  We selected the learning rate from the values [$9e-5$, $5e-5$, $1e-5$]. 

We performed all experiments with the same random seed, 76, used by invoking the \texttt{seed\_everything} function in PyTorch Lightning. 

All reported results are from a single model run. 

\subsection{Baselines} \label{appx:baselines}

\paragraph{\vertserini} For prediction on \scifact, we use the checkpoint available at \url{https://github.com/castorini/pygaggle/tree/master/experiments/vert5erini}. For \covidFact and \healthVer, we follow the instructions in that repository to convert the data to the required format, and train using the available training code as-is, beginning from the available \scifact checkpoint. We used Google Cloud TPU for training and inference. 

\paragraph{\pjoint} We use the code available at \url{https://github.com/jacklxc/ParagraphJointModel}. For predictions on \scifact, we make predictions using the publicly available checkpoint. For the other two target datasets, we use the training code in the repo without modification.

We used \pjoint as our baseline for zero / few-shot learning experiments, and hence also performed pretraining on \pjoint. The repository provides code to train on the \fever dataset, which we used for pretraining with \ei and \pqa added to the data.  

\section{Additional results and analysis} \label{appx:results}

\subsection{Full ablation results}

In Table \ref{tbl:ablations}, we presented F1 scores for ablations comparing pretraining data, model architecture, and encoder used. Table \ref{tbl:ablations_full} presents the full results, including precision and recall.

\begin{table*}[t]
    \setlength{\tabcolsep}{0.3em}
    \footnotesize
    \centering

    \begin{subtable}[h]{\linewidth}
        \begin{tabular}{
    p{0.05\linewidth}
    p{0.12\linewidth}
    P{0.026\linewidth}P{0.026\linewidth}P{0.033\linewidth}  P{0.026\linewidth}P{0.026\linewidth}P{0.033\linewidth} P{0.004\linewidth}
    P{0.026\linewidth}P{0.026\linewidth}P{0.033\linewidth}  P{0.026\linewidth}P{0.026\linewidth}P{0.033\linewidth} P{0.004\linewidth}
    P{0.026\linewidth}P{0.026\linewidth}P{0.033\linewidth}  P{0.026\linewidth}P{0.026\linewidth}P{0.033\linewidth} P{0.004\linewidth}
    }
    \toprule
                          & {}                   & \multicolumn{6}{c}{\textbf{\healthVer}}                                       &  & \multicolumn{6}{c}{\textbf{\covidFact}}                                       &  & \multicolumn{6}{c}{\textbf{\scifact}}                                        & \\
                          & {}                   & \multicolumn{3}{c}{Abstract}          & \multicolumn{3}{c}{Sentence}          &  & \multicolumn{3}{c}{Abstract}          & \multicolumn{3}{c}{Sentence}          &  & \multicolumn{3}{c}{Abstract}          & \multicolumn{3}{c}{Sentence}           \\
    \cmidrule(lr){3-5} \cmidrule(lr){6-8} \cmidrule(lr){10-12} \cmidrule(lr){13-15} \cmidrule(lr){17-19} \cmidrule(lr){20-22}
                          & \textbf{Pretraining} & \sz{P}    & \sz{R}    & F1            & \sz{P}    & \sz{R}    & F1            &  & \sz{P}    & \sz{R}    & F1            & \sz{P}    & \sz{R}    & F1            &  & \sz{P}    & \sz{R}    & F1            & \sz{P}    & \sz{R}    & F1             \\

    \midrule
    \multirow{2}{*}{Zero} & \feverSci            & \sz{60.6} & \sz{20.5} & \textbf{30.7} & \sz{25.0} & \sz{4.6}  & \textbf{7.8}  &  & \sz{48.8} & \sz{45.7} & \textbf{47.2} & \sz{32.7} & \sz{18.5} & \textbf{23.6} &  & \sz{49.0} & \sz{44.6} & \textbf{46.7} & \sz{39.0} & \sz{21.6} & \textbf{27.8}  \\
                          & \fever               & \sz{80.0} & \sz{0.7}  & 1.3           & \sz{66.7} & \sz{0.4}  & 0.7           &  & \sz{95.8} & \sz{14.5} & 25.2          & \sz{63.5} & \sz{6.2}  & 11.2          &  & \sz{83.8} & \sz{14.0} & 23.9          & \sz{64.9} & \sz{6.5}  & 11.8           \\
    \cmidrule(lr){1-22}
    \multirow{3}{*}{Few}  & \feverSci            & \sz{63.6} & \sz{47.9} & \textbf{54.7} & \sz{41.9} & \sz{31.0} & \textbf{35.7} &  & \sz{71.3} & \sz{68.1} & 69.7          & \sz{39.5} & \sz{35.4} & 37.4          &  & \sz{76.4} & \sz{54.1} & \textbf{63.3} & \sz{51.7} & \sz{40.3} & \textbf{45.3}  \\
                          & \fever               & \sz{56.4} & \sz{50.8} & 53.4          & \sz{34.8} & \sz{29.4} & 31.9          &  & \sz{74.4} & \sz{74.4} & \textbf{74.4} & \sz{39.3} & \sz{45.3} & \textbf{42.1} &  & \sz{72.4} & \sz{43.7} & 54.5          & \sz{48.8} & \sz{32.4} & 39.0           \\
                          & \targetOnly          & \sz{38.5} & \sz{40.4} & 39.4          & \sz{28.5} & \sz{25.7} & 27.0          &  & \sz{67.8} & \sz{67.8} & 67.8          & \sz{24.9} & \sz{20.7} & 22.6          &  & \sz{20.0} & \sz{30.6} & 24.2          & \sz{9.5}  & \sz{12.7} & 10.8           \\
    \cmidrule(lr){1-22}
    \multirow{3}{*}{Full} & \feverSci            & \sz{78.9} & \sz{76.3} & \textbf{77.6} & \sz{71.4} & \sz{67.0} & 69.1          &  & \sz{77.3} & \sz{77.3} & 77.3          & \sz{41.5} & \sz{46.1} & \textbf{43.7} &  & \sz{73.8} & \sz{71.2} & \textbf{72.5} & \sz{67.4} & \sz{67.0} & \textbf{67.2}  \\
                          & \fever               & \sz{77.5} & \sz{76.6} & 77.1          & \sz{70.8} & \sz{69.8} & \textbf{70.3} &  & \sz{77.5} & \sz{77.3} & \textbf{77.4} & \sz{40.6} & \sz{46.5} & 43.3          &  & \sz{64.3} & \sz{72.1} & 67.9          & \sz{57.1} & \sz{67.0} & 61.7           \\
                          & \targetOnly          & \sz{75.0} & \sz{74.0} & 74.5          & \sz{71.8} & \sz{67.8} & 69.7          &  & \sz{69.7} & \sz{69.7} & 69.7          & \sz{35.3} & \sz{38.1} & 36.6          &  & \sz{64.9} & \sz{61.7} & 63.3          & \sz{62.7} & \sz{54.6} & 58.4           \\
    \bottomrule
\end{tabular}

        \subcaption{Effect of pretraining data.}
        \label{tbl:ablations_pretrain_full}
    \end{subtable}

    \begin{subtable}[h]{\linewidth}
        \begin{tabular}{
    p{0.05\linewidth}
    p{0.12\linewidth}
    P{0.026\linewidth}P{0.026\linewidth}P{0.033\linewidth}  P{0.026\linewidth}P{0.026\linewidth}P{0.033\linewidth} P{0.004\linewidth}
    P{0.026\linewidth}P{0.026\linewidth}P{0.033\linewidth}  P{0.026\linewidth}P{0.026\linewidth}P{0.033\linewidth} P{0.004\linewidth}
    P{0.026\linewidth}P{0.026\linewidth}P{0.033\linewidth}  P{0.026\linewidth}P{0.026\linewidth}P{0.033\linewidth} P{0.004\linewidth}
    }
    \toprule
                          & {}               & \multicolumn{6}{c}{\textbf{\healthVer}}                                       &  & \multicolumn{6}{c}{\textbf{\covidFact}}                                       &  & \multicolumn{6}{c}{\textbf{\scifact}}                                        & \\
                          & {}               & \multicolumn{3}{c}{Abstract}          & \multicolumn{3}{c}{Sentence}          &  & \multicolumn{3}{c}{Abstract}          & \multicolumn{3}{c}{Sentence}          &  & \multicolumn{3}{c}{Abstract}          & \multicolumn{3}{c}{Sentence}           \\
    \cmidrule(lr){3-5} \cmidrule(lr){6-8} \cmidrule(lr){10-12} \cmidrule(lr){13-15} \cmidrule(lr){17-19} \cmidrule(lr){20-22}
                          & \textbf{Encoder} & \sz{P}    & \sz{R}    & F1            & \sz{P}    & \sz{R}    & F1            &  & \sz{P}    & \sz{R}    & F1            & \sz{P}    & \sz{R}    & F1            &  & \sz{P}    & \sz{R}    & F1            & \sz{P}    & \sz{R}    & F1             \\
    \midrule
    \multirow{2}{*}{Zero} & Longformer       & \sz{60.6} & \sz{20.5} & 30.7          & \sz{25.0} & \sz{4.6}  & 7.8           &  & \sz{48.8} & \sz{45.7} & 47.2          & \sz{32.7} & \sz{18.5} & 23.6          &  & \sz{49.0} & \sz{44.6} & \textbf{46.7} & \sz{39.0} & \sz{21.6} & \textbf{27.8}  \\
                          & RoBERTa          & \sz{59.5} & \sz{24.0} & \textbf{34.2} & \sz{25.4} & \sz{5.6}  & \textbf{9.2}  &  & \sz{49.3} & \sz{47.3} & \textbf{48.3} & \sz{35.2} & \sz{20.9} & \textbf{26.2} &  & \sz{45.5} & \sz{45.0} & 45.2          & \sz{34.4} & \sz{20.8} & 25.9           \\
    \cmidrule(lr){1-22}
    \multirow{2}{*}{Few}  & Longformer       & \sz{63.6} & \sz{47.9} & \textbf{54.7} & \sz{41.9} & \sz{31.0} & 35.7          &  & \sz{71.3} & \sz{68.1} & 69.7          & \sz{39.5} & \sz{35.4} & 37.4          &  & \sz{76.4} & \sz{54.1} & \textbf{63.3} & \sz{51.7} & \sz{40.3} & \textbf{45.3}  \\
                          & RoBERTa          & \sz{55.0} & \sz{47.9} & 51.2          & \sz{39.0} & \sz{35.0} & \textbf{36.9} &  & \sz{72.5} & \sz{71.6} & \textbf{72.1} & \sz{39.7} & \sz{42.5} & \textbf{41.0} &  & \sz{59.0} & \sz{44.1} & 50.5          & \sz{36.8} & \sz{31.6} & 34.0           \\
    \cmidrule(lr){1-22}
    \multirow{2}{*}{Full} & Longformer       & \sz{78.9} & \sz{76.3} & 77.6          & \sz{71.4} & \sz{67.0} & 69.1          &  & \sz{77.3} & \sz{77.3} & 77.3          & \sz{41.5} & \sz{46.1} & \textbf{43.7} &  & \sz{73.8} & \sz{71.2} & \textbf{72.5} & \sz{67.4} & \sz{67.0} & \textbf{67.2}  \\
                          & RoBERTa          & \sz{77.8} & \sz{80.0} & \textbf{78.8} & \sz{73.4} & \sz{72.0} & \textbf{72.7} &  & \sz{78.2} & \sz{78.2} & \textbf{78.2} & \sz{40.8} & \sz{46.3} & 43.4          &  & \sz{67.1} & \sz{68.0} & 67.6          & \sz{62.7} & \sz{61.9} & 62.3           \\
    \bottomrule
\end{tabular}

        \subcaption{Effect of base encoder.}
        \label{tbl:ablations_encoder_full}
    \end{subtable}

    \begin{subtable}[h]{\linewidth}
        \begin{tabular}{
    p{0.05\linewidth}
    p{0.12\linewidth}
    P{0.026\linewidth}P{0.026\linewidth}P{0.033\linewidth}  P{0.026\linewidth}P{0.026\linewidth}P{0.033\linewidth} P{0.004\linewidth}
    P{0.026\linewidth}P{0.026\linewidth}P{0.033\linewidth}  P{0.026\linewidth}P{0.026\linewidth}P{0.033\linewidth} P{0.004\linewidth}
    P{0.026\linewidth}P{0.026\linewidth}P{0.033\linewidth}  P{0.026\linewidth}P{0.026\linewidth}P{0.033\linewidth} P{0.004\linewidth}
    }
    \toprule
                          & {}                & \multicolumn{6}{c}{\textbf{\healthVer}}                                       &  & \multicolumn{6}{c}{\textbf{\covidFact}}                                       &  & \multicolumn{6}{c}{\textbf{\scifact}}                                        & \\
                          & {}                & \multicolumn{3}{c}{Abstract}          & \multicolumn{3}{c}{Sentence}          &  & \multicolumn{3}{c}{Abstract}          & \multicolumn{3}{c}{Sentence}          &  & \multicolumn{3}{c}{Abstract}          & \multicolumn{3}{c}{Sentence}           \\
    \cmidrule(lr){3-5} \cmidrule(lr){6-8} \cmidrule(lr){10-12} \cmidrule(lr){13-15} \cmidrule(lr){17-19} \cmidrule(lr){20-22}
    \textbf{}             & \textbf{Approach} & \sz{P}    & \sz{R}    & F1            & \sz{P}    & \sz{R}    & F1            &  & \sz{P}    & \sz{R}    & F1            & \sz{P}    & \sz{R}    & F1            &  & \sz{P}    & \sz{R}    & F1            & \sz{P}    & \sz{R}    & F1             \\
    \midrule
    \multirow{3}{*}{Zero} & Multitask         & \sz{60.6} & \sz{20.5} & \textbf{30.7} & \sz{25.0} & \sz{4.6}  & \textbf{7.8}  &  & \sz{48.8} & \sz{45.7} & \textbf{47.2} & \sz{32.7} & \sz{18.5} & \textbf{23.6} &  & \sz{49.0} & \sz{44.6} & \textbf{46.7} & \sz{39.0} & \sz{21.6} & \textbf{27.8}  \\
                          & Pipe              & \sz{58.8} & \sz{1.7}  & 3.2           & \sz{29.4} & \sz{0.5}  & 0.9           &  & \sz{67.3} & \sz{11.0} & 19            & \sz{57.4} & \sz{5.8}  & 10.5          &  & \sz{80.6} & \sz{13.1} & 22.5          & \sz{72.2} & \sz{7.0}  & 12.8           \\
                          & MT / PI           & \sz{60.9} & \sz{2.3}  & 4.5           & \sz{41.7} & \sz{0.9}  & 1.8           &  & \sz{78.5} & \sz{16.1} & 26.7          & \sz{57.7} & \sz{7.6}  & 13.5          &  & \sz{80.9} & \sz{17.1} & 28.3          & \sz{75.5} & \sz{10.0} & 17.7           \\
    \cmidrule(lr){1-22}
    \multirow{3}{*}{Few}  & Multitask         & \sz{63.6} & \sz{47.9} & \textbf{54.7} & \sz{41.9} & \sz{31.0} & \textbf{35.7} &  & \sz{71.3} & \sz{68.1} & \textbf{69.7} & \sz{39.5} & \sz{35.4} & 37.4          &  & \sz{76.4} & \sz{54.1} & \textbf{63.3} & \sz{51.7} & \sz{40.3} & \textbf{45.3}  \\
                          & Pipe              & \sz{56.3} & \sz{49.7} & 52.8          & \sz{32.6} & \sz{27.0} & 29.5          &  & \sz{69.4} & \sz{67.2} & 68.3          & \sz{40.6} & \sz{36.0} & \textbf{38.2} &  & \sz{54.8} & \sz{51.4} & 53.0          & \sz{43.7} & \sz{36.8} & 39.9           \\
                          & MT / PI           & \sz{67.0} & \sz{35.9} & 46.7          & \sz{44.5} & \sz{25.3} & 32.3          &  & \sz{72.6} & \sz{50.2} & 59.3          & \sz{40.2} & \sz{29.7} & 34.1          &  & \sz{85.3} & \sz{41.9} & 56.2          & \sz{54.7} & \sz{33.0} & 41.1           \\
    \cmidrule(lr){1-22}
    \multirow{3}{*}{Full} & Multitask         & \sz{78.9} & \sz{76.3} & 77.6          & \sz{71.4} & \sz{67.0} & 69.1          &  & \sz{77.3} & \sz{77.3} & 77.3          & \sz{41.5} & \sz{46.1} & 43.7          &  & \sz{73.8} & \sz{71.2} & \textbf{72.5} & \sz{67.4} & \sz{67.0} & \textbf{67.2}  \\
                          & Pipe              & \sz{78.7} & \sz{78.1} & \textbf{78.4} & \sz{70.2} & \sz{68.3} & \textbf{69.2} &  & \sz{79.9} & \sz{75.4} & \textbf{77.6} & \sz{48.2} & \sz{47.2} & \textbf{47.7} &  & \sz{68.5} & \sz{73.4} & 70.9          & \sz{64.5} & \sz{68.1} & 66.2           \\
                          & MT / PI           & \sz{77.6} & \sz{64.8} & 70.6          & \sz{70.0} & \sz{59.5} & 64.3          &  & \sz{77.7} & \sz{69.4} & 73.3          & \sz{43.6} & \sz{44.4} & 44.0          &  & \sz{80.5} & \sz{48.2} & 60.3          & \sz{70.5} & \sz{47.8} & 57.0           \\
    \bottomrule
\end{tabular}

        \subcaption{Effect of model architecture.}
        \label{tbl:ablations_architecture_full}
    \end{subtable}

    \caption{
        Full ablation results.
    }
    \label{tbl:ablations_full}
\end{table*}

\subsection{Performance on rationales with undefined acronyms} \label{appx:acronyms}

In \S \ref{sec:analysis_pipeline}, we examined the difference in performance on instances with self-contained vs. context-dependent evidence. Here, we show the results of evaluation on instances containing an undefined acronym vs. cases without one. We find that undefined acronyms do not pose a challenge for Multitask and Pipeline, but do cause a small performance drop on MT / PI.

\begin{table}[t]
    \setlength{\tabcolsep}{0.3em}
    \footnotesize
    \centering

    \begin{tabular}{
        L{4.5em}
        *{3}{P{1.3em}}
        P{0.5em}
        *{3}{P{1.3em}}
        P{0.5em}
        P{3em}
    }
    \toprule
    {}                & \multicolumn{3}{c}{\tworowsc{\textbf{No undefined} \\ \textbf{acronym}}} &  & \multicolumn{3}{c}{\tworowsc{\textbf{Undefined} \\ \textbf{acronym}}} &  &                      \\
    \\
    \cmidrule(lr){2-4}  \cmidrule(lr){6-8}
    \textbf{Approach} & \sz{P}    & \sz{R}    & F1                                               &  & \sz{P}    & \sz{R}    & F1                                            &  & $\mathbf{\% \Delta}$ \\
    \midrule
    Multitask         & \sz{88.1} & \sz{73.8} & 80.3                                             &  & \sz{86.0} & \sz{77.1} & 81.3                                          &  & 1.2\%                \\
    Pipeline          & \sz{89.9} & \sz{77.5} & 83.2                                             &  & \sz{88.6} & \sz{81.2} & 84.8                                          &  & 1.9\%                \\
    MT / PI           & \sz{97.1} & \sz{42.5} & 59.1                                             &  & \sz{85.0} & \sz{35.4} & 50.0                                          &  & -15.4\%              \\
    \cmidrule(lr){1-10}
    Count             & \multicolumn{3}{l}{80}                                                   &  & \multicolumn{3}{l}{48}                                                &  &                      \\
    \bottomrule
\end{tabular}

    \caption{
    Performance of different modeling approaches on instances with vs. without an undefined acronym. We perform evaluation on the same data as reported in Table \ref{tbl:skills_context}.
    }
    \label{tbl:skills_acronym}
\end{table}

\subsection{Negative sampling} \label{appx:negative_sampling}

In \S \ref{sec:model_training} we described how, for \scifact, we trained on 20 negative abstracts per claim. The effect of training on these additional negative samples is shown in Figure \ref{tbl:negative_sampling}. In the abstract-provided setting, negative sampling is not very beneficial. However, when the model must select evidence from retrieved abstracts, precision drops off dramatically without negative sampling. This is worth noting since it suggests that performance reported when models are provided with ``gold'' candidate abstracts may not offer an accurate estimate of the accuracy these systems would achieve when deployed in a real-world setting, which could require systems to verify claims over hundreds of thousands of documents.

\begin{table}[t]
  \footnotesize
  \setlength{\tabcolsep}{0.3em}
  \centering

  \begin{tabular}{
    p{0.17\linewidth}
    p{0.17\linewidth}
    P{0.07\linewidth}P{0.07\linewidth}P{0.09\linewidth}  P{0.07\linewidth}P{0.07\linewidth}P{0.09\linewidth}
    }
    \toprule
    \tworows{Retrieval}             & \tworows{Neg.                                           \\ sample} & \multicolumn{3}{c}{Abstract}          & \multicolumn{3}{c}{Sentence}          \\
    \cmidrule(lr) {3-5} \cmidrule(lr) {6-8}

                                    &                                                                    & \sz{P}    & \sz{R}    & F1            & \sz{P}    & \sz{R}    & F1            \\
    \midrule
    \tworows{Abstract- \\ provided} & \xmark                                                             & \sz{81.9} & \sz{85.6} & \textbf{83.7} & \sz{69.5} & \sz{69.7} & 69.6          \\
                                    & \cmark                                                             & \sz{85.2} & \sz{75.2} & 79.9          & \sz{79.0} & \sz{70.3} & \textbf{74.4} \\
    \cmidrule(lr){1-8}
    \multirow{2}{*}{Open}           & \xmark                                                             & \sz{38.9} & \sz{80.6} & 52.5          & \sz{35.4} & \sz{65.1} & 45.9          \\
                                    & \cmark                                                             & \sz{73.8} & \sz{71.2} & \textbf{72.5} & \sz{67.4} & \sz{67.0} & \textbf{67.2} \\
    \bottomrule
\end{tabular}

  \caption{Effect of negative sampling on \scifact.}

  \label{tbl:negative_sampling}

\end{table}

\subsection{Cross-dataset generalization} \label{appx:generalization}

\begin{table}[t]
  \footnotesize
  \setlength{\tabcolsep}{0.4em}
  \centering

  \begin{subtable}[h]{\linewidth}
    \centering
    \begin{tabular}{
  p{0.21\linewidth}
  R{0.09\linewidth}R{0.09\linewidth}P{0.0005\linewidth}
  R{0.09\linewidth}R{0.09\linewidth}P{0.0005\linewidth}
  R{0.09\linewidth}R{0.09\linewidth}P{0.0005\linewidth}
  }
  \toprule
  Eval $\rightarrow$ & \multicolumn{2}{c}{\healthVer} &          & \multicolumn{2}{c}{\covidFact} &      & \multicolumn{2}{c}{\scifact}                      \\
  \cmidrule(lr){2-3} \cmidrule(lr){5-6} \cmidrule(lr){8-9}
  Train $\downarrow$ & F1                             & $\Delta$ &                                & F1   & $\Delta$                     &  & F1   & $\Delta$ \\
  \midrule
  \healthVer         & 86.1                           & 0.0      &                                & 50.2 & -24.0                        &  & 73.4 & -15.8    \\
  \covidFact         & 50.6                           & -35.6    &                                & 74.1 & 0.0                          &  & 76.1 & -13.1    \\
  \scifact           & 70.5                           & -15.7    &                                & 54.6 & -19.6                        &  & 89.2 & 0.0      \\
  \cmidrule(lr){1-9}
  Combined           & 83.0                           & -3.2     &                                & 64.3 & -9.8                         &  & 87.8 & -1.3     \\
  \bottomrule
\end{tabular}

    \subcaption{Abstract-level evaluation. \scifact and \healthVer generalize fairly well to each other. \covidFact generalizes well to \scifact, but not \healthVer.}
    \label{tbl:cross_train_abstract}
  \end{subtable}
  \hspace{0.4em}
  \begin{subtable}[h]{\linewidth}
    \centering
    \begin{tabular}{
  p{0.21\linewidth}
  R{0.09\linewidth}R{0.09\linewidth}P{0.0005\linewidth}
  R{0.09\linewidth}R{0.09\linewidth}P{0.0005\linewidth}
  R{0.09\linewidth}R{0.09\linewidth}P{0.0005\linewidth}
  }
  \toprule
  Eval $\rightarrow$ & \multicolumn{2}{c}{\healthVer} &          & \multicolumn{2}{c}{\covidFact} &      & \multicolumn{2}{c}{\scifact}                      \\
  \cmidrule(lr){2-3} \cmidrule(lr){5-6} \cmidrule(lr){8-9}
  Train $\downarrow$ & F1                             & $\Delta$ &                                & F1   & $\Delta$                     &  & F1   & $\Delta$ \\
  \midrule
  \healthVer         & 74.2                           & 0.0      &                                & 28.0 & -12.6                        &  & 39.7 & -32.4    \\
  \covidFact         & 14.6                           & -59.5    &                                & 40.6 & 0.0                          &  & 41.6 & -30.6    \\
  \scifact           & 20.5                           & -53.7    &                                & 33.9 & -6.7                         &  & 72.1 & 0.0      \\
  \cmidrule(lr){1-9}
  Combined           & 71.4                           & -2.8     &                                & 39.8 & -0.9                         &  & 70.5 & -1.6     \\
  \bottomrule
\end{tabular}

    \subcaption{Sentence-level evaluation. None of the datasets generalize particularly well to each other. \healthVer generalizes better to \scifact than vice versa.}
    \label{tbl:cross_train_sentence}
  \end{subtable}

  \caption{Cross-dataset generalization performance. The rows and columns indicate the training and evaluation datasets, respectively. The $\Delta$ values indicate the loss in performance from evaluating on a dataset different from the one the model was trained on. The ``Combined'' row indicates training on all datasets combined.}
  \label{tbl:cross_train}

\end{table}

In \S \ref{sec:experimental_setup}, we discussed how the available scientific fact-checking datasets differ in a number of important respects. Here, we explore whether models trained on one system are able to generalize to another despite these differences.  We train \sysname on each of our three datasets and then evaluate its performance on the other two. We also train a version of \sysname on all three datasets together, and evaluate on each one. Since \covidFact has no \nei instances, during evaluation we remove all \nei instances from the other two datasets, and evaluate in the abstract-provided setting.

The results are shown in Table \ref{tbl:cross_train}. The sentence-level evaluation results (Table \ref{tbl:cross_train_sentence}) indicate that none of the datasets generalize well to each other in their ability to identify rationales. The situation is better for abstract labeling (Table \ref{tbl:cross_train_abstract}). \scifact and \healthVer each generalize reasonably well to each other, but not to \covidFact. \covidFact generalizes well to \scifact, but not to \healthVer. In general, \scifact appears the ``easiest'' dataset to generalize to; this could be explained by the fact that \scifact claims were written to be atomic and therefore simple to verify.

Finally, a model trained on all datasets combined manages to achieve reasonable performance across all three datasets, though falling short of the performance of models trained specifically for each individual dataset.

\end{document}